\documentclass[journal,twoside]{IEEEtran}

\usepackage{cite}
\usepackage{amsmath,amssymb,amsfonts}
\usepackage{graphicx}
\usepackage{booktabs}
\usepackage{textcomp}
\usepackage{algorithm}
\usepackage{algorithmic}
\usepackage[hidelinks]{hyperref}

\makeatletter
\renewcommand{\IEEEauthorrefmark}[1]{\textsuperscript{\footnotesize #1}}
\makeatother

\begin{document}

\title{SwinTF3D: A Lightweight Multimodal Fusion Approach for Text-Guided 3D Medical Image Segmentation}

\author{
\IEEEauthorblockN{
Hasan~Faraz~Khan\IEEEauthorrefmark{1},
Noor~Fatima\IEEEauthorrefmark{1},
Muzammil~Behzad\IEEEauthorrefmark{1}\IEEEauthorrefmark{2}\IEEEauthorrefmark{*}
}
\\
\IEEEauthorblockA{\IEEEauthorrefmark{1}King Fahd University of Petroleum and Minerals, Saudi Arabia}
\\
\IEEEauthorblockA{\IEEEauthorrefmark{2}SDAIA-KFUPM Joint Research Center for Artificial Intelligence, Saudi Arabia}
\\
\IEEEauthorblockA{Emails: g202427420@kfupm.edu.sa, g202427440@kfupm.edu.sa, muzammil.behzad@kfupm.edu.sa}
\thanks{\IEEEauthorrefmark{*}Corresponding author: Muzammil Behzad.}
}

\maketitle

\begin{abstract}
The recent integration of artificial intelligence into medical imaging has driven remarkable advances in automated organ segmentation. However, most existing 3D segmentation frameworks rely exclusively on visual learning from large annotated datasets restricting their adaptability to new domains and clinical tasks. The lack of semantic understanding in these models makes them ineffective in addressing flexible, user-defined segmentation objectives. To overcome these limitations, we propose SwinTF3D, a lightweight multimodal fusion approach that unifies visual and linguistic representations for text-guided 3D medical image segmentation. The model employs a transformer-based visual encoder to extract volumetric features and integrates them with a compact text encoder via an efficient fusion mechanism. This design allows the system to understand natural-language prompts and correctly align semantic cues with their corresponding spatial structures in medical volumes, while producing accurate, context-aware segmentation results with low computational overhead. Extensive experiments on the BTCV dataset demonstrate that SwinTF3D achieves competitive Dice and IoU scores across multiple organs, despite its compact architecture. The model generalizes well to unseen data and offers significant efficiency gains compared to conventional transformer-based segmentation networks. Bridging visual perception with linguistic understanding, SwinTF3D establishes a practical and interpretable paradigm for interactive, text-driven 3D medical image segmentation, opening perspectives for more adaptive and resource-efficient solutions in clinical imaging.
\end{abstract}



\begin{IEEEkeywords}
Medical Image Segmentation, Vision–Language Models, Text-Guided Segmentation, Multimodal Fusion, Transformer-Based Architecture
\end{IEEEkeywords}

\section{Introduction}
Medical image segmentation plays a vital role in modern computer-assisted diagnosis, treatment planning, and surgical navigation. It allows precise delineation of anatomical structures from complex imaging modalities such as computed tomography (CT) and magnetic resonance imaging (MRI), providing the spatial and morphological information necessary for accurate clinical decision-making. Among the different anatomical regions, abdominal organ segmentation is one of the most challenging and clinically significant tasks. It contributes to several applications, including liver volumetry, kidney function assessment, tumor detection, and radiotherapy planning. However, achieving accurate segmentation of multiple abdominal organs is still difficult because of the large variability among patients, overlapping organ boundaries, low soft-tissue contrast in CT scans, and variations in imaging protocols across different scanners and institutions  \cite{Tadokoro2023, Wang2022}.

Over the past few years, the development of deep learning has dramatically advanced the field of medical image segmentation, particularly in three-dimensional (3D) imaging. Traditional convolutional neural networks (CNNs) and encoder–decoder designs have long served as the foundation for automated segmentation, consistently outperforming region-based or intensity-driven techniques \cite{Chen2021a}. Later, transformer-based architectures were introduced, which greatly improved the ability to capture long-range spatial dependencies and contextual relationships. Several studies have shown that these transformer-based networks outperform conventional models in complex multi-organ segmentation tasks by modeling both global and local information effectively \cite{Cao2021, Zhou2023}. Although such models have achieved excellent accuracy, most of them focus entirely on visual processing and rely heavily on large volumes of annotated data \cite{chang2024emnetefficientchannel}.

Researchers have recently started investigating multimodal learning strategies that integrate textual and visual data in order to get around these limitations. However, bringing such multimodal frameworks into medical imaging remains a demanding challenge. Earlier studies that attempted text-guided segmentation often required large-scale pretraining on multimodal datasets containing millions of image–text pairs \cite{Chen2024, Li20241}. Such resources are rarely available in medical imaging, where text data such as radiology reports are typically unstructured and differ substantially from the dense, voxel-level supervision required for segmentation \cite{Xin2025}. Moreover, these large multimodal systems tend to be computationally expensive, making them impractical for fine-tuning on small, domain-specific datasets. As a result, their ability to generalize across new organs, modalities, or patient populations is often limited. In the field of 3D medical segmentation, several studies have also explored more efficient alternatives. Some have focused on simplifying model complexity, while others have aimed to integrate textual information through lighter fusion mechanisms instead of using heavy cross-attention decoders. A few recent works have demonstrated that fusion performed directly in the logit space can achieve strong results with much lower computational cost \cite{zhao2025, liang2023}. Such approaches use fusion techniques to combine textual and visual representations rather than adding multiple trainable transformer layers.

In this study, we present a lightweight multimodal approach called SwinTF3D for text-guided 3D abdominal organ segmentation. The model is designed to integrate both visual and linguistic understanding while remaining computationally efficient. Our experiments are conducted on the BTCV (Beyond the Cranial Vault) dataset, a widely used benchmark for multi-organ abdominal segmentation that contains thirteen different abdominal organs \cite{harrigr2015}. This dataset provides a suitable environment to test how well the model generalizes across structures of varying size, appearance, and spatial arrangement. The proposed approach not only achieves accurate 3D segmentation of multiple abdominal organs but also allows the process to be directed through natural language descriptions.

\begin{figure}[t]
\centering
\includegraphics[width=\columnwidth]{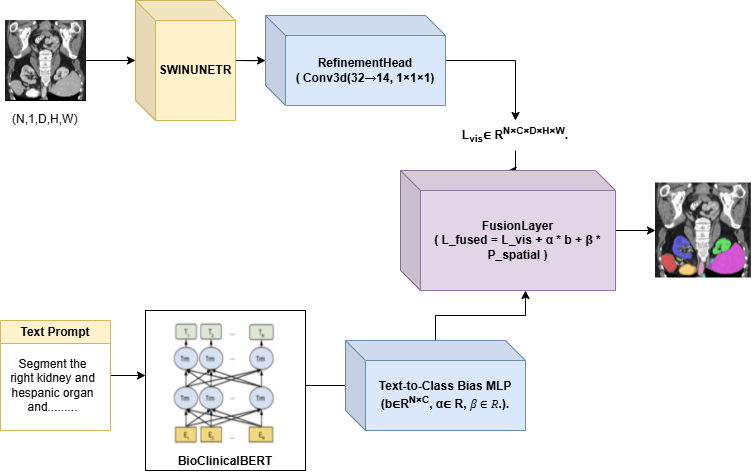}
\caption{Overall architecture of the proposed SwinTF3D model showing the image and text processing pathways, the fusion module, and the final segmentation output.}
\label{fig:swintext3d_architecture}
\end{figure}

The overall design of the proposed model is illustrated in Fig.~\ref{fig:swintext3d_architecture}, which shows how SwinTF3D integrates both image and text modalities through a lightweight fusion mechanism to generate text-guided 3D segmentation outputs.

The main objective of this work is to create a model capable of understanding both the spatial characteristics of 3D medical images and the semantic meaning contained in textual prompts. To accomplish this, we designed a dual-branch structure consisting of a visual encoder and a language encoder, which are linked through a fusion module. The visual encoder is based on a transformer-driven design that captures volumetric and anatomical details from CT scans. We fine-tuned this encoder on the BTCV dataset, adding a lightweight 3D refinement head to improve its adaptation to abdominal CT characteristics and organ boundaries. This fine-tuned encoder serves as the foundation of our multimodal system. Simultaneously, the text encoder interprets natural language prompts and transforms them into semantic embeddings that reflect the purpose of each instruction \cite{He2023}. A lightweight mathematical mechanism that modifies the prediction based on the linguistic input is then used to fuse these embeddings with the visual logits \cite{liang2023}. This fusion approach maintains accuracy and flexibility while drastically lowering computational demand by avoiding a large multimodal decoder.
Conceptually, our approach transforms traditional segmentation into a task-driven process where textual input dynamically influences the model’s predictions. Rather than simply assigning fixed anatomical labels, the system can interpret references to specific organs or spatial relations and adapt its focus accordingly. This semantic conditioning allows the system to handle diverse segmentation tasks without retraining, while remaining efficient and interpretable.

Large-scale multimodal systems are powerful but resource-heavy and data-dependent. In contrast, our framework leverages compact encoders and efficient training strategies, enabling practical deployment even on limited hardware. By fine-tuning a pre-existing Swin-based 3D segmentation model as the visual backbone and combining it with a small language model for text understanding, we achieve a balance between performance, flexibility, and scalability \cite{Hatamizadeh2022}. The fusion mechanism, which is implemented through a small set of learnable layers, acts as a bridge between modalities, learning to correlate visual features with linguistic representations during training. Once trained, the entire model can accept both a 3D image and a textual prompt to produce the corresponding segmentation volume.

Experimentally, we demonstrated the effectiveness of the proposed approach on the BTCV dataset for multi-organ abdominal segmentation. The model achieves high Dice and Intersection-over-Union scores across multiple organs, validating its ability to segment complex anatomical structures in 3D volumes. Moreover its text-guided inference capability allows users to specify the organs of interest dynamically, offering an additional level of control absent in conventional segmentation pipelines. These results indicate that the proposed lightweight multimodal approach can serve as a practical and interpretable alternative to large-scale vision–language segmentation frameworks particularly in medical imaging contexts where data and computational resources are constrained.

\begin{figure}[t]
    \centering
    \includegraphics[width=\columnwidth]{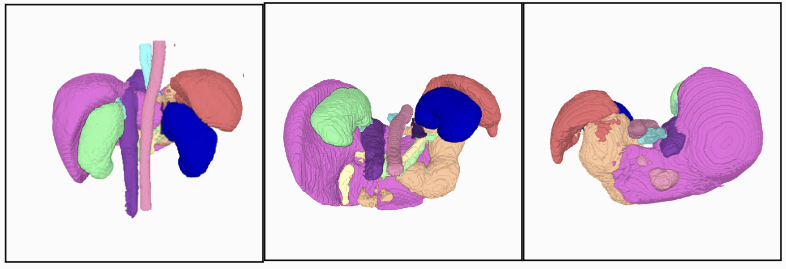}
    \caption{3D visualization of abdominal organ segmentation results on the BTCV dataset. The reconstructed surfaces are obtained from the predicted segmentation masks generated by SwinTF3D, highlighting accurate anatomical delineation and spatial consistency across multiple organs.}
    \label{fig:3d_visual_results}
\end{figure}

The 3D visualization of the segmented abdominal organs obtained is shown in Fig.~\ref{fig:3d_visual_results}. These visuals demonstrate how the predicted segmentation aligns with anatomical structures in three dimensions, providing an intuitive representation of organ shape, orientation, and spatial relationships.

In summary, the contributions of this work are both methodological and practical. Methodologically, we introduce an efficient multimodal approach that fuses text and visual information for 3D medical image segmentation. From an application standpoint, we establish a working prototype for text-driven multi-organ segmentation, showing that semantic guidance can greatly improve both the flexibility and usability of medical segmentation systems. Overall, the goal of SwinTF3D is to make segmentation not only more accurate but also more interactive, interpretable, and aligned with real-world clinical reasoning.

\section{Related Work}
\subsection{Transformer-Based Visual Encoders in Medical Image Segmentation.}
The introduction of transformer-based architectures brought a significant improvement in medical image segmentation by enabling models to capture long-range spatial dependencies that traditional convolutional neural networks (CNNs) often fail to model effectively. Early frameworks such as TransUNet \cite{Chen2021b} combined convolutional backbones with Vision Transformers and created a hybrid encoder–decoder design that incorporated both local texture details and global contextual information. Building on this foundation SwinUNETR extended these principles to volumetric segmentation tasks using a hierarchical Swin Transformer encoder with shifted-window self-attention. This approach effectively modeled the 3D spatial context and made it one of the most widely adopted backbones in transformer-based medical segmentation \cite{He2023, Hatamizadeh2022}.  
Subsequent research focused on improving hierarchical feature representation and computational efficiency. nnFormer \cite{Zhou2023} introduced a multi-scale token aggregation mechanism, integrating local convolutions and global attention to achieve better context modeling across scales. Similarly, MISSFormer \cite{Huang2019, Huang2021} employed a multi-scale attention design that adaptively balanced local and global feature extraction which improved organ boundary delineation and segmentation accuracy. The UNETR++ model refined this family of transformer-based architectures through progressive skip connections and re-designed decoder pathways, addressing feature redundancy and improving gradient flow during optimization \cite{Isensee2024, Huang2019}. Collectively, these models established the foundation for 3D segmentation encoders, providing high-quality volumetric feature extraction and robust anatomical representation, which serve as the visual backbone in many subsequent multimodal frameworks, including ours.

\subsection{Text-Guided and Multimodal Segmentation.}
Although transformer-based architectures have advanced visual representation, they remain limited to static visual cues and cannot adapt dynamically to context-dependent clinical queries. To address this limitation, researchers have explored multimodal integration, particularly vision–language alignment, to enable models that can respond to natural-language prompts. LViT (Language meets Vision Transformer) proposed a unified transformer that processes both text and image tokens, achieving early-stage alignment between linguistic semantics and visual features \cite{Li20242}. Extending this concept to volumetric imaging, Med3DVLM developed a vision–language model capable of understanding 3D medical data through large language model embeddings that guide segmentation via text conditioning \cite{Xin2025}.  
The rise of text-guided segmentation frameworks has further broadened this multimodal field. A research adapted the Segment Anything Model (SAM) for volumetric segmentation, demonstrating that prompt-driven segmentation can generalize to 3D medical data without explicit retraining \cite{Xie2024}. Similarly, SAT (Segment Anything in 3D Medical Images with Text Prompts) introduced a large-vocabulary segmentation framework where text prompts define the segmentation targets, bridging open-vocabulary image segmentation and medical imaging. Dynamic prompt generation further improved this paradigm by generating task-specific textual embeddings directly from the visual input, allowing dynamic adaptation to new organ combinations\cite{Ndir2025, chen2023gen}. MedUniSeg unified prompt-based segmentation across both 2D and 3D modalities, showcasing how text conditioning can generalize across imaging domains while maintaining strong segmentation accuracy.  
Together, these works demonstrate a shift from static label-based segmentation toward language-driven medical AI systems that are more interactive, interpretable, and aligned with clinical reasoning.

\subsection{Fusion-Based and Decoder-Free Multimodal Approaches.}
While text-guided models enable flexible segmentation, many existing approaches rely on heavy multimodal decoders or cross-attention mechanisms, which increase computational overhead. To overcome these challenges, several studies have explored more efficient fusion mechanisms that operate directly in feature or logit space rather than through complex decoders. The E2ENet model proposed an end-to-end multimodal fusion strategy where textual cues were aligned with multi-view visual features without the need for a large decoding head. Similarly, TextDiffSeg employed a diffusion-based latent-space fusion mechanism between text and visual embeddings, performing prompt-driven segmentation with a smaller computational footprint \cite{Liu2024, Ma2025}.  
In a similar direction, the Mask-Adapted CLIP (OVSeg) model from open-vocabulary segmentation fused text embeddings with visual logits using a linear combination, producing effective segmentation masks without requiring a trainable decoder \cite{liang2023}. This concept of operating directly in logit space through mathematical formulations rather than deep feature decoders has influenced many efficient multimodal frameworks in both natural and medical imaging. In the medical domain, lightweight architectures such as TK-Mamba and MISSFormer also emphasized feature refinement through token routing and selective fusion rather than heavy attention-based modules \cite{Wang2023, Li20243, Das2025}.  
These methods collectively illustrate that text–vision fusion can be realized through parameter-efficient mathematical operations that directly modify class probabilities or feature logits, a concept that closely aligns with the fusion mechanism employed in this work.

Overall, the literature in multimodal medical image segmentation can be divided into two main streams. The first comprises transformer-based encoders such as TransUNet, SwinUNETR, nnFormer, MISSFormer, and UNETR++, which specialize in volumetric representation learning but remain limited to visual information. The second stream introduces text guidance through vision–language modeling, represented by LViT, Med3DVLM, Text3DSAM, SAT, Dynamic Prompt Generation, and MedUniSeg, where textual input provides semantic control for segmentation tasks. More recent fusion-based studies such as E2ENet, TextDiffSeg, and Mask-Adapted CLIP explore mathematical fusion techniques in logit space, replacing heavy decoders with lightweight additive or scaling operations.  
Building upon these developments, our proposed SwinTF3D model unifies the strengths of hierarchical 3D transformer encoding and efficient multimodal fusion. Instead of adding complex attention-based decoders, it fuses textual and visual cues directly in the logit space through a parameter-efficient biasing mechanism, enabling prompt-driven 3D organ segmentation that is both computationally lightweight and semantically interpretable.

\section{Methodology}

Recent advances in multimodal medical segmentation have demonstrated that coupling volumetric encoders with language-based conditioning can substantially enhance the interpretability and flexibility of segmentation systems. However, most existing studies achieve this by training large-scale vision–language models from scratch or through extensive pretraining on tens of thousands of volumetric samples. Similarly, some researchs explored prompt-driven segmentation but employed heavy optimization pipelines and large-scale text–image pairings to achieve generalization across organs and modalities \cite{Xie2024, Li2025}. These approaches, while effective, highlight two major challenges that persist in the field. First, large multimodal medical models often depend on resource-intensive pretraining, which limits their accessibility and reproducibility in realistic research or clinical settings. Second, existing datasets such as BTCV and Synapse lack aligned linguistic descriptions, making text–mask alignment difficult without additional data engineering or external supervision.

The objective of this work is to develop a more training-efficient multimodal segmentation framework that leverages pretrained, domain-specific encoders rather than end-to-end large-scale retraining. By fine-tuning a strong volumetric backbone (Swin-UNETR) and introducing a lightweight fusion mechanism for text–vision integration, the proposed SwinTF3D framework aims to achieve prompt-based three-dimensional segmentation while maintaining computational feasibility. This approach reduces dependence on large training corpora and enables flexible, clinically interpretable segmentation through natural language guidance.

We adopt a pretrained Swin-UNETR as the volumetric visual encoder and introduce a compact fusion head that maps a frozen biomedical text embedding to class-wise logit biases, optionally modulated by relation-aware spatial priors derived from the prompt. The visual backbone remains frozen for fusion training, preserving its anatomical representations while the multimodal head learns how text should modulate the segmentation logits. This design emphasizes modularity, training efficiency, and practical deployability. The proposed fine-tuned Swin-UNETR model with an additional refinement head is referred to as SwinUNETR-RH, while the overall multimodal fusion framework proposed in this work is named SwinTF3D.


The proposed SwinTF3D framework performs text-guided three-dimensional segmentation of abdominal organs. The task can be formally defined as a conditional mapping problem. Given a volumetric CT image $I \in \mathbb{R}^{H \times W \times D}$ and a textual prompt $T$, the goal is to learn a function
\begin{equation}
f_{\theta}(I, T) = S,
\end{equation}
where $S \in \mathbb{R}^{H \times W \times D}$ is the predicted segmentation mask and $\theta$ denotes the model parameters. The text input $T$ specifies the target organ(s), thereby conditioning the segmentation output on a linguistic description rather than a fixed label index.

Let $\mathcal{C} = \{c_0, c_1, ..., c_{13}\}$ be the set of organ classes, where $c_0$ denotes the background class. For each voxel $v_i$ in the image, the model outputs a class probability vector $p_i = [p_i(c_0), p_i(c_1), \ldots, p_i(c_{13})]$ such that
\begin{equation}
\sum_{c=0}^{13} p_i(c) = 1, \qquad p_i(c) \geq 0.
\end{equation}
The final segmentation map $S$ is generated by taking the class with the maximum posterior probability at each voxel:
\begin{equation}
S(v_i) = \arg\max_{c \in \mathcal{C}} p_i(c).
\end{equation}
This probabilistic formulation allows the model to reason about uncertainty across organs and align segmentation predictions with the semantics encoded in the text.

\subsection{Visual Encoder: The Swin-UNETR Architecture}

The visual encoder in our framework is based on the Swin-UNETR architecture, a transformer-based model designed for volumetric medical image segmentation. It combines convolutional neural networks (CNNs) for local feature extraction with vision transformers for modeling long-range dependencies, enabling the capture of both global and fine-grained anatomical patterns \cite{Chen2023, Cao2021}.

The input CT volume $I \in \mathbb{R}^{H \times W \times D}$ is partitioned into non-overlapping 3D patches of size $4 \times 4 \times 4$. Each patch is flattened and projected into a $C$-dimensional embedding space:
\begin{equation}
z_0 = [x_1E; x_2E; \ldots; x_N E] + E_{\text{pos}},
\end{equation}
where $E$ is the learnable embedding matrix, $E_{\text{pos}}$ is the positional encoding, and $N = \frac{HWD}{4^3}$.

Unlike standard transformers, Swin-UNETR restricts self-attention to local 3D windows to reduce computational cost while preserving contextual awareness. The window-based self-attention is computed as:
\begin{equation}
\text{Attention}(Q, K, V) = \text{Softmax}\left(\frac{QK^{\top}}{\sqrt{d_k}} + B \right)V,
\end{equation}
where $B$ denotes the relative positional bias. To enable cross-window interaction, windows are shifted by $\frac{M}{2}$ voxels in alternating layers, allowing efficient context aggregation without global attention overhead \cite{Wang2023, Cao2021}.

The encoder comprises four hierarchical stages with Swin Transformer blocks and patch merging layers, producing multi-scale feature maps $\{F_1, F_2, F_3, F_4\}$:
\begin{equation}
F_k = \text{SwinBlock}_k(F_{k-1}), \quad k = 1, 2, 3, 4.
\end{equation}
The decoder mirrors this hierarchy, progressively upsampling features and integrating them via skip connections. The final segmentation logits are obtained as:
\begin{equation}
\hat{S} = D(F_1, F_2, F_3, F_4),
\end{equation}
where $D(\cdot)$ denotes the decoding function.

Skip connections fuse encoder and decoder features as
\begin{equation}
\tilde{F}_k = \text{Concat}(F_k, U_{k+1}),
\end{equation}
followed by convolutional refinement to preserve spatial coherence.

Overall, Swin-UNETR balances local detail preservation and global context modeling through hierarchical attention and skip connections, making it well suited for accurate 3D medical image segmentation.

\subsection{Fine-Tuning of the Visual Encoder}

Although pretrained transformer-based models provide strong initialization, domain-specific fine-tuning is crucial for achieving optimal performance in medical imaging tasks. In this work, we propose an architecture that fine-tunes the baseline Swin-UNETR model and introduces a lightweight refinement head appended to the output of the pretrained backbone. This fine-tuned version is referred to as SwinUNETR-RH, where RH denotes the Refinement Head. The model was fine-tuned using the BTCV abdominal dataset to adapt it to the intensity profiles, organ shapes, and contextual relationships unique to abdominal CT imaging  \cite{Shaker2022}.

To further enhance segmentation quality, we introduced a lightweight 3D refinement head appended to the output of the pretrained backbone. The updated refinement head employs \emph{instance normalization} and \emph{ReLU} activation with a small dropout, providing stable gradients during training. It consists of a $3\times3\times3$ convolution, normalization, activation, dropout, and a final $1\times1\times1$ projection, combined with a residual connection. The refinement operation can be formulated as:
\begin{equation}
R(x) = \text{Conv}_{1\times1\times1}\!\left(\text{Dropout}\!\left(\text{ReLU}\!\left(\text{IN}\!\left(\text{Conv}_{3\times3\times3}(x)\right)\right)\right)\right),
\end{equation}
and the refined logits are computed as
\begin{equation}
\tilde{S} = S + R(S).
\end{equation}

For optimization, we employed a hybrid loss based on the Dice–Focal formulation, which effectively balances region overlap and class imbalance, especially for smaller anatomical structures \cite{Taha2015, Wang2022}. The Dice–Focal loss can be expressed as:
\begin{equation}
\mathcal{L}_{\text{total}} = \mathcal{L}_{\text{Dice}} + \mathcal{L}_{\text{Focal}},
\end{equation}
where the Dice loss is defined as
\begin{equation}
\mathcal{L}_{\text{Dice}} = 1 - \frac{2 \sum_i p_i g_i + \epsilon}{\sum_i p_i^2 + \sum_i g_i^2 + \epsilon},
\end{equation}
and the focal component is given by
\begin{equation}
\mathcal{L}_{\text{Focal}} = -\frac{1}{N}\sum_{i=1}^{N}\sum_{c=0}^{13}(1-p_i(c))^{\gamma}\,g_i(c)\log(p_i(c)),
\end{equation}
where $\gamma=2.0$. The model was optimized with AdamW (two learning-rate groups) and cosine annealing, in mixed precision.

\begin{figure*}[t]
    \centering
    \includegraphics[width=\textwidth]{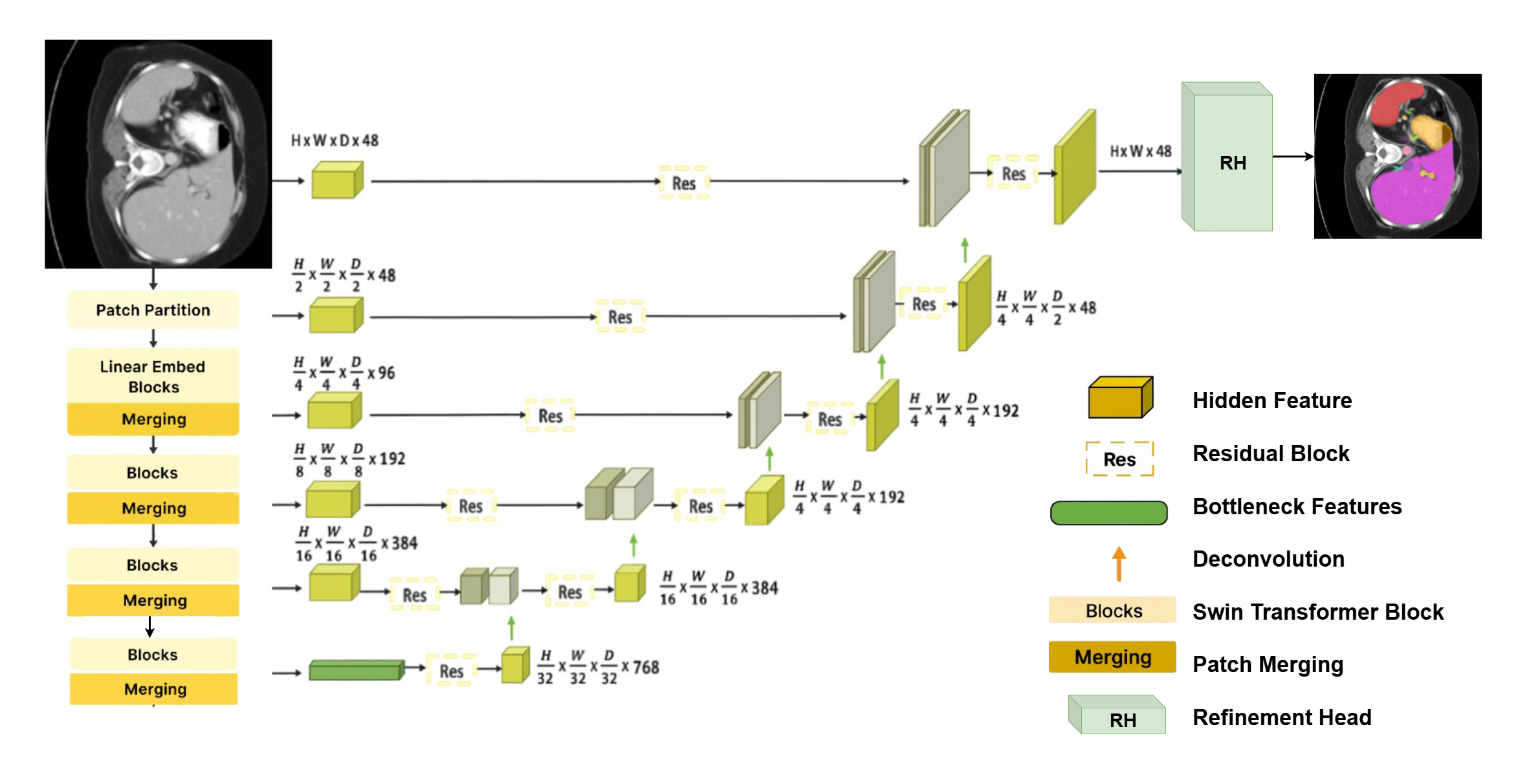}
    \caption{Architecture of the SwinUNETR-RH model used as the visual encoder. The network employs hierarchical Swin Transformer blocks for volumetric feature extraction and incorporates a lightweight refinement head (RH) for improved boundary precision in organ segmentation.}
    \label{fig:swinunetr_rh_arch}
\end{figure*}

Performance during fine-tuning was monitored using the Dice Similarity Coefficient (DSC) and the Intersection-over-Union (IoU):
\begin{equation}
\text{DSC} = \frac{2|P \cap G|}{|P| + |G|}, \quad 
\text{IoU} = \frac{|P \cap G|}{|P \cup G|}.
\end{equation}
The fine-tuned SwinUNETR-RH augmented with the refined 3D head, serves as the visual encoder for the proposed SwinTF3D framework. The detailed architecture of the SwinUNETR-RH visual encoder used in the proposed SwinTF3D framework is shown in Fig.~\ref{fig:swinunetr_rh_arch}. The model consists of hierarchical Swin Transformer blocks for volumetric feature extraction, followed by a lightweight refinement head (RH) that enhances boundary precision in the segmentation output. Each stage of the encoder captures multi-scale contextual features, while the refinement head refines the decoded representation to produce sharper organ boundaries and improved segmentation accuracy \cite{Hu2025},.

\subsection{Text Encoder}

The linguistic component of our framework is designed to interpret descriptive text prompts and encode them into continuous semantic embeddings. We adopt a frozen biomedical language model, \emph{BioClinicalBERT}, which provides domain-aligned language representations \cite{sounack2025}. This encoder converts each textual prompt $T$ into a dense vector $t \in \mathbb{R}^{E}$, where $E = 768$ denotes the embedding dimension:
\begin{equation}
t = \phi_{\text{text}}(T).
\end{equation}
During both training and inference, the text encoder remains frozen, ensuring stability and preventing overfitting to the relatively small set of textual prompts used in this work. Given a batch of $N$ prompts,
\begin{equation}
T_{\text{emb}} = [t_1, t_2, \ldots, t_N] \in \mathbb{R}^{N \times E}.
\end{equation}
To detect which anatomical classes are explicitly mentioned in the prompt, we employ a text–pattern matching function $\psi(T)$ that maps organ names and synonyms to class identifiers, producing a binary presence vector $y^{\text{text}} \in \{0,1\}^{C}$.

\subsection{Multimodal Fusion Mechanism}

To align textual semantics with volumetric visual features, we introduce a fusion module that injects text-derived biases into the 3D segmentation logits produced by the Swin-UNETR encoder. The fusion component is realized through a lightweight multilayer perceptron, referred to as the \textit{Text-to-Class Bias} block, which maps the textual embedding to class-wise bias coefficients:
\begin{equation}
b = W_2 \, \text{ReLU}(W_1 t + b_1) + b_2, \quad b \in \mathbb{R}^{C},
\end{equation}
where $W_1 \in \mathbb{R}^{E \times H}$, $W_2 \in \mathbb{R}^{H \times C}$, and $H=256$. Two learnable scalars, $\alpha$ and $\beta$, control the global influence of text bias and spatial priors on the final segmentation. The combined logits are
\begin{equation}
L_{\text{fused}} = L_{\text{vis}} + \alpha \, b[:, :, \text{None}, \text{None}, \text{None}] + \beta \, P_{\text{spatial}},
\end{equation}
where $L_{\text{vis}} \in \mathbb{R}^{N \times C \times D \times H \times W}$ are the visual logits and $P_{\text{spatial}}$ is the spatial prior tensor.

\paragraph{Spatial Priors and Anatomical Relations.}
The model incorporates optional spatial priors to capture anatomical dependencies such as “the region around the kidney that belongs to the liver.” Given a predicted segmentation $S_{\text{pred}}$ and anchor class $a$, the prior for a target class $t$ is computed using dilation and an Euclidean distance transform:
\begin{equation}
P_{\text{spatial}}(v) = \max\left(0,\, 1 - \frac{d(v, \Omega_a)}{d_{\text{max}} + 1}\right),
\end{equation}
where $d(v, \Omega_a)$ is the distance to the anchor region $\Omega_a$ and $d_{\text{max}}$ controls the dilation extent.

\subsection{Training Strategy}

To align textual semantics with volumetric visual features, we introduce a fusion module that injects text-derived biases into the 3D segmentation logits produced by the SwinUNETR-RH encoder. This fusion module is implemented as a lightweight multilayer perceptron, termed the \textit{Text-to-Class Bias} block, which maps textual embeddings to class-wise bias coefficients. The overall training objective is defined as:
\begin{equation}
\mathcal{L}_{\text{total}} = \mathcal{L}_{\text{seg}} + \lambda_{\text{text}} \, \mathcal{L}_{\text{text}} + \lambda_{\text{rel}} \, \mathcal{L}_{\text{rel}},
\end{equation}
where $\mathcal{L}_{\text{seg}}$ is the volumetric segmentation loss, $\mathcal{L}_{\text{text}}$ enforces alignment between textual intent and predicted organ presence, and $\mathcal{L}_{\text{rel}}$ regularizes spatial relations between organs.

\paragraph{Segmentation Loss.}
The segmentation loss $\mathcal{L}_{\text{seg}}$ uses the Dice–Cross Entropy hybrid formulation:
\begin{equation}
\mathcal{L}_{\text{seg}} = \mathcal{L}_{\text{Dice}} + \mathcal{L}_{\text{CE}},
\end{equation}
where both components are computed on the fused logits $L_{\text{fused}}$ and corresponding ground-truth masks.

\paragraph{Text Alignment Loss.}
To ensure that textual prompts influence only the referenced organs, a binary cross-entropy loss is applied between the predicted class biases and the organ presence vector:
\begin{equation}
\mathcal{L}_{\text{text}} 
= -\frac{1}{C}\sum_{c=1}^{C} 
\big[y_c^{\text{text}}\log(\sigma_c) + (1 - y_c^{\text{text}})\log(1 - \sigma_c)\big],
\end{equation}
where $\sigma(\cdot)$ denotes the sigmoid activation. This loss aligns textual semantics with organ-specific activations.

\paragraph{Relation-Aware Loss.}
For prompts containing relational expressions, spatial priors are introduced to encourage coherent multi-organ segmentation. For each relation $(a,t)$, a masked cross-entropy loss is computed within the corresponding spatial region:
\begin{equation}
\mathcal{L}_{\text{rel}} = \frac{1}{R}\sum_{r=1}^{R} 
\frac{\sum_{v \in \Omega_r} \text{CE}(L_{\text{fused}}(v), G_t(v))}{
|\Omega_r|}
\end{equation}
where $\Omega_r$ denotes the spatial prior region for relation $r$, $R$ is the number of relations in the batch, and $G_t$ is the target organ mask. This term enforces smooth transitions across relational boundaries.

Training uses the AdamW optimizer with a learning rate of $2\times10^{-3}$, weight decay $1\times10^{-4}$, and cosine annealing over 20 epochs. Only the fusion parameters are updated, while the SwinUNETR-RH encoder and text transformer remain frozen. The overall training procedure is summarized in Algorithm~\ref{alg:fusion-training}.

\begin{algorithm}[ht]
\caption{Training Procedure for SwinTF3D}
\label{alg:fusion-training}
\begin{algorithmic}[1]
\REQUIRE Training dataset $\{(I_n, G_n)\}_{n=1}^{N}$, corresponding text prompts $\{T_n\}_{n=1}^{N}$
\FOR{epoch = 1 to $E$}
    \FOR{each mini-batch $(I, G, T)$}
        \STATE Encode the text prompts: $t \leftarrow \phi_{\text{text}}(T)$
        \STATE Compute class bias and scaling factors: $(b, \alpha, \beta) \leftarrow \phi_{\text{fusion}}(t)$
        \STATE Generate visual logits from the fine-tuned encoder: $L_{\text{vis}} \leftarrow f_{\text{visual}}(I)$
        \STATE Fuse text and visual representations:
        \begin{equation*}
        L_{\text{fused}} \leftarrow L_{\text{vis}} + \alpha b + \beta P_{\text{spatial}}
        \end{equation*}
        \STATE Compute segmentation loss $\mathcal{L}_{\text{seg}}$, text alignment loss $\mathcal{L}_{\text{text}}$ and relation-aware loss $\mathcal{L}_{\text{rel}}$
        \STATE Combine all losses:
        \begin{equation*}
        \mathcal{L}_{\text{total}} = \mathcal{L}_{\text{seg}} + 
        \lambda_{\text{text}}\mathcal{L}_{\text{text}} + 
        \lambda_{\text{rel}}\mathcal{L}_{\text{rel}}
        \end{equation*}
        \STATE Update fusion model parameters by minimizing $\mathcal{L}_{\text{total}}$
    \ENDFOR
    \STATE Adjust learning rate using cosine annealing schedule
\ENDFOR
\RETURN Trained fusion parameters $\theta_{\text{fusion}}$
\end{algorithmic}
\end{algorithm}

\subsection{Inference and Text-Guided Segmentation}

During inference, the model accepts a volumetric CT scan $I$ and a textual prompt $T$. The prompt is parsed to identify the set of mentioned organs and any inter-organ relations. The embedding $t = \phi_{\text{text}}(T)$ is used to compute the class bias $b$, and the visual encoder generates segmentation logits $L_{\text{vis}}$. These are fused using the trained parameters $\alpha$ and $\beta$:
\begin{equation}
L_{\text{fused}} = L_{\text{vis}} + \alpha \, b + \beta \, P_{\text{spatial}}.
\end{equation}
The final segmentation mask is obtained by applying the softmax operator over classes and taking the voxel-wise argmax:
\begin{equation}
S(v_i) = \arg\max_{c} \, \text{Softmax}(L_{\text{fused}})_c(v_i).
\end{equation}
This mechanism allows flexible segmentation driven by natural-language instructions.

\section{Experiments and Results}
This section presents the experimental evaluation of the proposed SwinTF3D framework. The experiments are designed to validate the effectiveness of multimodal fusion for text-guided three-dimensional medical image segmentation and to compare the performance of the fine-tuned visual encoder (SwinUNETR-RH) with the full text-vision fusion model. All experiments were implemented using the MONAI and PyTorch frameworks, and the results are reported on volumetric CT datasets with standardized preprocessing and synthetic prompt construction.

\subsection{Datasets and Preprocessing}

\subsubsection{Datasets}
The experiments were primarily conducted on the Beyond the Cranial Vault (BTCV) multi-organ abdominal CT dataset, which serves as a widely used benchmark for 3D medical segmentation. The BTCV dataset consists of 50 volumetric contrast-enhanced CT scans, each with voxel-level annotations for 13 abdominal organs: spleen, right kidney, left kidney, gallbladder, esophagus, liver, stomach, aorta, inferior vena cava, portal and splenic vein, pancreas, right adrenal gland, and left adrenal gland. The dataset captures significant anatomical and intensity variations arising from differences in scanner types, acquisition protocols, and patient demographics, providing a rigorous evaluation scenario for volumetric segmentation models. For additional qualitative validation, the Synapse multi-organ CT dataset was used to evaluate generalization capability. Synapse includes 30 CT scans annotated for 8 organs and provides complementary anatomical coverage to BTCV. The model was trained on BTCV and qualitatively tested on Synapse without additional fine-tuning to demonstrate cross-dataset transferability.

\subsubsection{Preprocessing}
All preprocessing was implemented using the MONAI library to ensure consistency and reproducibility. Each CT volume was loaded in NIfTI format using the NibabelReader interface. The preprocessing pipeline standardized orientation, resolution, and intensity across the dataset.

All volumes were reoriented to the RAS (Right–Anterior–Superior) coordinate system using Orientationd and resampled to an isotropic spacing of $1.5 \times 1.5 \times 2.0~\text{mm}$ via Spacingd, with spline interpolation for images and nearest-neighbor interpolation for labels. Intensities were clipped to the Hounsfield Unit range $[-175, 250]$ and normalized to $[0,1]$ using ScaleIntensityRanged. Background regions were removed using CropForegroundd, and the volumes were divided into fixed-size patches of $96 \times 96 \times 96$ voxels using RandCropByPosNegLabeld to balance organ-containing and background samples.

During training, data augmentation included random flipping along all axes, random 90° rotations, and random intensity shifts up to 10\%. All samples were converted to PyTorch tensors using EnsureTyped. To improve efficiency, a caching strategy was applied using CacheDataset with cache sizes of 8 for training and 4 for validation.

\subsubsection{Synthetic Prompt Generation}
As the BTCV and Synapse datasets lack natural language or PROM-style annotations, a synthetic prompt corpus was generated to train the text-guided segmentation module. A total of 650 prompts were used for training and 130 for validation and inference. Prompts were automatically constructed using organ labels, anatomical groupings, and curated synonym and relation templates.

Prompt generation followed a probabilistic strategy in which 1–3 organs were randomly selected per volume to form instructions such as “segment the liver and spleenic organ” or “segment the right kidney and the hepatic organ.” Relational phrases were occasionally included to simulate anchor–target relationships, and synonym substitutions (e.g., “hepatic organ,” “renal structure”) were used to increase lexical diversity. Each prompt was aligned with its corresponding image volume by matching mentioned organs to the label maps, and each training iteration paired a 3D CT patch with a randomly sampled prompt describing organs present in that patch.

\subsection{Evaluation Metrics}

To comprehensively evaluate the segmentation performance of the proposed SwinTF3D framework, multiple quantitative metrics were employed. These metrics capture the overlap, accuracy, and boundary consistency between the predicted segmentation mask and the ground truth. Let $P$ denote the predicted voxel set, $G$ the ground-truth voxel set, and $C$ the total number of organ classes (including background).

\subsubsection{Dice Similarity Coefficient (DSC)}
The Dice Similarity Coefficient (DSC) quantifies the degree of overlap between prediction and ground truth and is defined as:
\begin{equation}
\text{DSC}(P,G) = \frac{2|P \cap G|}{|P| + |G|} = \frac{2\sum_{i=1}^{N} p_i g_i}{\sum_{i=1}^{N} p_i + \sum_{i=1}^{N} g_i},
\end{equation}
where $p_i, g_i \in \{0,1\}$ denote voxel-wise binary labels for prediction and ground truth, and $N$ is the total number of voxels in the 3D volume. The DSC ranges from 0 to 1, with higher values indicating better spatial overlap.

\subsubsection{Intersection over Union (IoU)}
The Intersection over Union (IoU), also known as the Jaccard Index, measures the intersection area relative to the union of the predicted and ground-truth masks:
\begin{equation}
\text{IoU}(P,G) = \frac{|P \cap G|}{|P \cup G|} = \frac{\sum_{i=1}^{N} p_i g_i}{\sum_{i=1}^{N} \big(p_i + g_i - p_i g_i\big)}.
\end{equation}
IoU penalizes both false positives and false negatives, providing a stricter measure of segmentation accuracy than DSC.

\subsubsection{Mean Intersection over Union (mIoU)}
For multi-organ segmentation, the mean IoU (mIoU) aggregates the IoU values across all foreground classes:
\begin{equation}
\text{mIoU} = \frac{1}{C-1}\sum_{c=1}^{C-1} \text{IoU}_c,
\end{equation}
where each $\text{IoU}_c$ corresponds to the IoU for organ class $c$ and the background class is excluded from averaging.

\subsubsection{Hausdorff Distance (HD$_{95}$)}
To assess boundary-level accuracy, the 95th percentile Hausdorff Distance (HD$_{95}$) is computed between the predicted and reference surfaces.  
Let $D_{PG}$ and $D_{GP}$ denote the directed distances from prediction to ground truth and vice versa:
\begin{align}
D_{PG} &= \max_{p \in P} \min_{g \in G} \|p - g\|_2, \\
D_{GP} &= \max_{g \in G} \min_{p \in P} \|g - p\|_2.
\end{align}
The 95th percentile of each directed distance is then taken to reduce the effect of outliers:
\begin{align}
d_{PG}^{95} &= \text{quantile}_{0.95}(D_{PG}), \\
d_{GP}^{95} &= \text{quantile}_{0.95}(D_{GP}).
\end{align}
Finally, the symmetric HD$_{95}$ is defined as:
\begin{equation}
\text{HD}_{95}(P,G) = \max(d_{PG}^{95},~ d_{GP}^{95}).
\end{equation}
A lower HD$_{95}$ value indicates tighter boundary alignment and fewer outlier deviations.

\subsubsection{Relative Volume Difference (RVD)}
The Relative Volume Difference (RVD) evaluates whether a model tends to over-segment or under-segment a structure:
\begin{equation}
\text{RVD}(P,G) = \frac{|P| - |G|}{|G|} \times 100\%.
\end{equation}
A perfect segmentation yields $\text{RVD} = 0$, while positive or negative values indicate over- and under-segmentation, respectively.

\subsubsection{Per-Organ Aggregation}
For each organ class $c \in \{1, \dots, C-1\}$, all metrics are computed independently and then averaged:
\begin{equation}
\mathcal{M}_{\text{avg}} = \frac{1}{C-1}\sum_{c=1}^{C-1}\mathcal{M}_c,
\end{equation}
where $\mathcal{M}_c$ can represent DSC, IoU, or HD$_{95}$ for organ $c$. This provides both per-organ and global performance insights.

\subsubsection{Inference Procedure}
During inference, the SwinTF3D model predicts segmentation masks conditioned on natural language prompts. Given an input volumetric CT scan $I \in \mathbb{R}^{H\times W\times D}$ and a text prompt $T$, the model first encodes $T$ using a frozen biomedical text encoder to obtain a text embedding $t = \phi_{\text{text}}(T)$. The SwinUNETR-RH visual encoder produces visual logits $L_{\text{vis}} = f_{\text{visual}}(I)$. A class-wise bias vector $b$ and scaling factors $\alpha$ and $\beta$ are computed as:
\begin{equation}
(b, \alpha, \beta) = \phi_{\text{fusion}}(t),
\end{equation}
and the final fused logits are computed by:
\begin{equation}
L_{\text{fused}} = L_{\text{vis}} + \alpha b[:, :, \text{None}, \text{None}, \text{None}] + \beta P_{\text{spatial}},
\end{equation}
where $P_{\text{spatial}}$ represents the spatial prior tensor derived from relational patterns in the prompt.

The segmentation prediction is obtained via softmax normalization and voxel-wise argmax:
\begin{equation}
S(v_i) = \arg\max_{c \in \mathcal{C}} \text{Softmax}(L_{\text{fused}})_c(v_i).
\end{equation}

To formalize this inference mechanism, Algorithm~\ref{alg:inference} summarizes the forward procedure of SwinTF3D in a mathematical format.

\begin{algorithm}[ht]
\caption{Inference Procedure for SwinTF3D}
\label{alg:inference}
\begin{algorithmic}[1]
\REQUIRE Input CT volume $I \in \mathbb{R}^{H\times W\times D}$, text prompt $T$
\STATE Encode text prompt: $t \leftarrow \phi_{\text{text}}(T)$
\STATE Compute fusion parameters: $(b, \alpha, \beta) \leftarrow \phi_{\text{fusion}}(t)$
\STATE Obtain visual logits: $L_{\text{vis}} \leftarrow f_{\text{visual}}(I)$
\STATE If relations exist in $T$, compute spatial prior $P_{\text{spatial}}$ via:
\begin{equation*}
P_{\text{spatial}}(v) = \max\left(0, 1 - \frac{d(v, \Omega_a)}{d_{\max}+1}\right)
\end{equation*}
\STATE Fuse modalities:
\begin{equation*}
L_{\text{fused}} = L_{\text{vis}} + \alpha b[:, :, \text{None}, \text{None}, \text{None}] + \beta P_{\text{spatial}}
\end{equation*}
\STATE Apply softmax to fused logits:
\begin{equation*}
p_c(v) = \frac{\exp(L_{\text{fused},c}(v))}{\sum_{k=0}^{C-1}\exp(L_{\text{fused},k}(v))}
\end{equation*}
\STATE Generate final segmentation:
\begin{equation*}
S(v) = \arg\max_{c \in \mathcal{C}} p_c(v)
\end{equation*}
\RETURN Segmentation mask $S$
\end{algorithmic}
\end{algorithm}

This inference process enables SwinTF3D to perform context-aware 3D medical segmentation conditioned on free-text prompts, allowing organ-level and relational segmentation from a single unified model.

\subsection{Experimental Setup}

\subsubsection{Implementation Details}
All experiments were conducted using the PyTorch framework (version 2.1) with the MONAI medical imaging library (version 1.3) for data loading, preprocessing, and training. NVIDIA CUDA and cuDNN were used for GPU acceleration, and all experiments ran on a single NVIDIA Tesla T4 GPU with 16 GB VRAM.

To ensure reproducibility, random seeds were fixed across PyTorch, NumPy, and Python’s random module. Mixed-precision training (AMP) was enabled to reduce memory usage and improve efficiency. The codebase was implemented in Python 3.10 and executed on a Linux system with CUDA 12.2 and cuDNN 8.9. Visualization and evaluation plots were generated using Matplotlib and MONAI utilities.

\subsubsection{Hyperparameter Configuration}
The SwinTF3D framework comprises two components: (1) the fine-tuned SwinUNETR-RH visual encoder and (2) the multimodal fusion head. Both were trained under a consistent protocol to balance performance and computational efficiency.

The SwinUNETR backbone was initialized with pretrained weights and fine-tuned for 15 epochs using a batch size of 1 and an input patch size of $96\times96\times96$. The encoder learning rate was set to $1\times10^{-4}$, while the refinement head used $5\times10^{-4}$ with the AdamW optimizer and a weight decay of $1\times10^{-5}$. A cosine annealing scheduler with 10 cycles was applied, and a hybrid Dice–Focal loss was employed to address class imbalance.

For fusion training, only the text-to-class bias module was trainable, while the SwinUNETR-RH encoder and BioClinicalBERT text encoder were frozen. AdamW was used with a learning rate of $2\times10^{-3}$, weight decay $1\times10^{-4}$, and cosine annealing over 20 epochs. Each epoch sampled up to 100 iterations with randomly generated synthetic prompts. The total loss combined segmentation, text alignment, and relation-aware losses as
\begin{equation}
\mathcal{L}_{\text{total}} = \mathcal{L}_{\text{seg}} + \lambda_{\text{text}} \mathcal{L}_{\text{text}} + \lambda_{\text{rel}} \mathcal{L}_{\text{rel}},
\end{equation}
where $\lambda_{\text{text}}=0.2$ and $\lambda_{\text{rel}}=0.2$. Gradient accumulation and mixed precision were applied to maintain stable optimization under limited GPU memory.

\subsubsection{Baselines and Comparison Models}
To evaluate the fine-tuned SwinUNETR-RH encoder, we compared it with several state-of-the-art 3D segmentation models using identical preprocessing and evaluation settings. The baselines include U-Net, TransUNet, Swin-UNet, UNETR, Swin-UNETR, nnUNet, nnFormer, and UNETR++, covering both convolutional and transformer-based paradigms widely used for BTCV organ segmentation.

These models represent key approaches in 3D medical image segmentation, from classical CNN-based designs to modern hierarchical transformers. The proposed SwinUNETR-RH was evaluated under a \textit{single-model} setting without ensemble learning or cross-validation to ensure fair comparison \cite{Shaker2022,Huang2019}.

While some prior studies report higher Dice scores on BTCV, many rely on ensemble methods, multi-fold cross-validation, or extended training data \cite{Liu2025,Roy2023}. Since SwinUNETR-RH is trained and evaluated as a single model, comparisons in this work are restricted to single-model baselines to provide an equitable and consistent performance assessment.

\subsection{Results}

This section presents the quantitative and qualitative results of the proposed SwinUNETR-RH and SwinTF3D models. The performance was evaluated on the BTCV dataset, which serves as the primary benchmark, and additional results were reported on the Synapse dataset to verify cross-dataset generalization. The results are compared with several recent transformer-based and convolutional baselines to validate the effectiveness of the proposed approach.

\subsubsection{Results on BTCV Dataset (SwinUNETR-RH)}

\begin{figure}[t]
    \centering
    \includegraphics[width=\columnwidth]{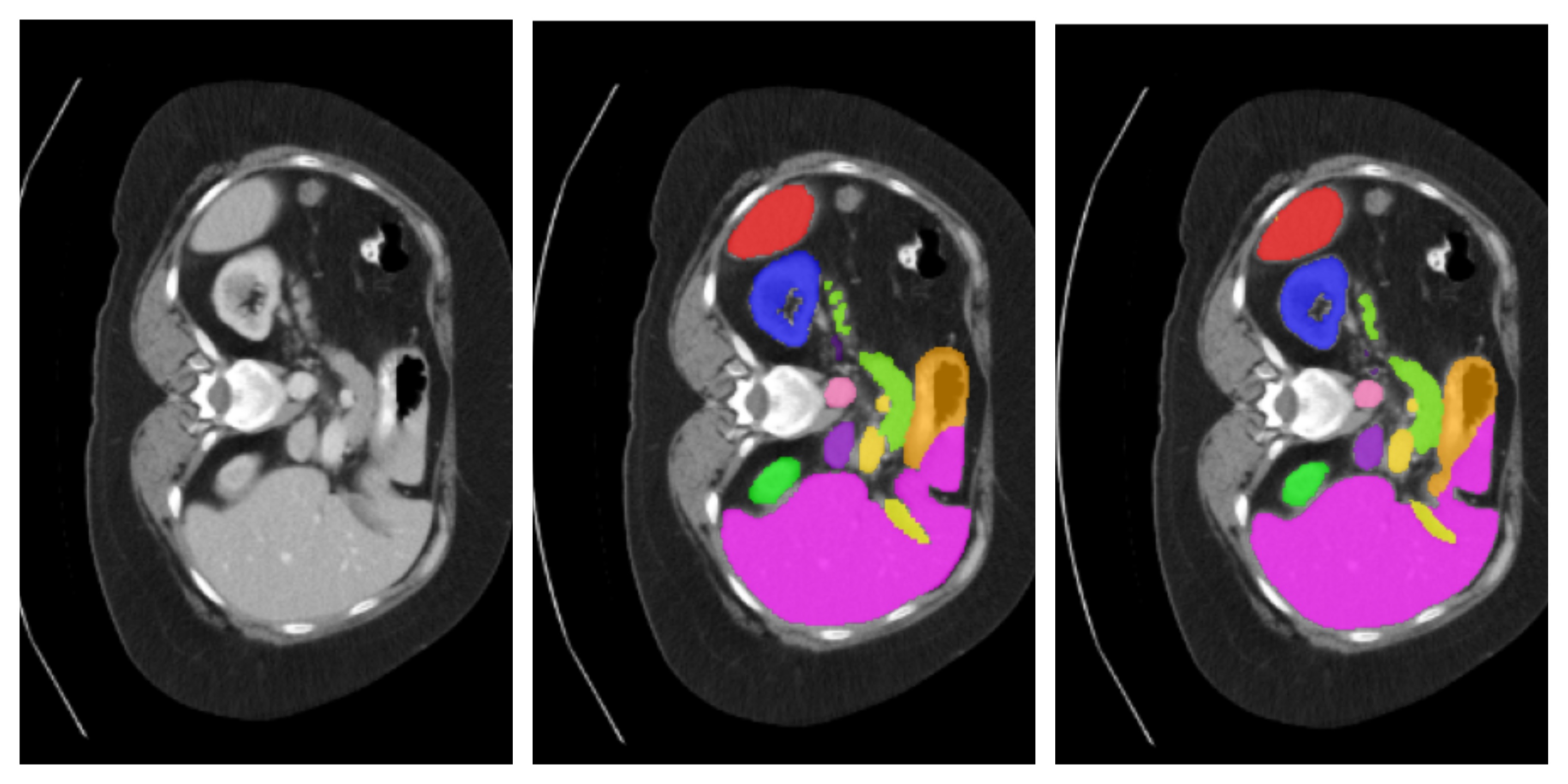}
    \caption{Axial view comparison of abdominal organ segmentation on the BTCV dataset. From left to right: the original CT slice, the ground truth segmentation, and the prediction generated by the proposed SwinTF3D framework.}
    \label{fig:axial_segmentation_results}
\end{figure}

Table~\ref{tab:organwise_metrics} summarizes the organ-wise quantitative results of the SwinUNETR-RH model on the BTCV dataset. The model achieves an average Dice Similarity Coefficient (DSC) of 0.8101 and a mean Intersection over Union (IoU) of 0.7057 across 13 abdominal organs. The precision and recall values remain well-balanced, reflecting stable performance across both large and small anatomical structures. The low Hausdorff Distance (HD95) scores for most organs indicate accurate boundary localization and minimal surface deviation from the ground truth. Fig.~\ref{fig:axial_segmentation_results}, demonstrates the qualitative axial view of multi-organ abdominal segmentation on the BTCV dataset. 

\begin{figure}[t]
    \centering
    \includegraphics[width=\columnwidth]{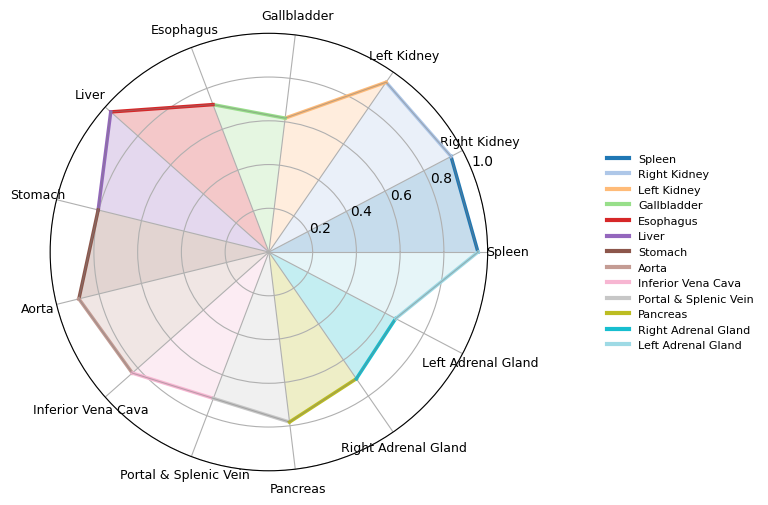}
    \caption{Average Dice Similarity Coefficient (DSC) per organ obtained using the proposed SwinTF3D framework on the BTCV dataset. The figure highlights the model’s consistent performance across multiple organs, with higher DSC values indicating more accurate segmentations.}
    \label{fig:dsc_per_organ}
\end{figure}

In particular, the SwinUNETR-RH model achieved high DSC scores for large and well-defined organs such as the liver (0.9659), spleen (0.9558), and kidneys (above 0.94), demonstrating robust volumetric representation learning. Smaller and anatomically complex organs such as the gallbladder, pancreas, and adrenal glands showed comparatively lower but consistent performance, which is expected given their small volume and irregular boundaries. These results confirm that fine-tuning the Swin-UNETR backbone with the refinement head significantly improves spatial precision and general segmentation reliability across varied organ sizes.As shown in Fig.~\ref{fig:dsc_per_organ}, the average Dice Similarity Coefficient (DSC) values are reported for each organ, providing a comparative overview of segmentation performance across all anatomical structures.

\begin{figure}[!htbp]
    \centering
    \includegraphics[width=\columnwidth]{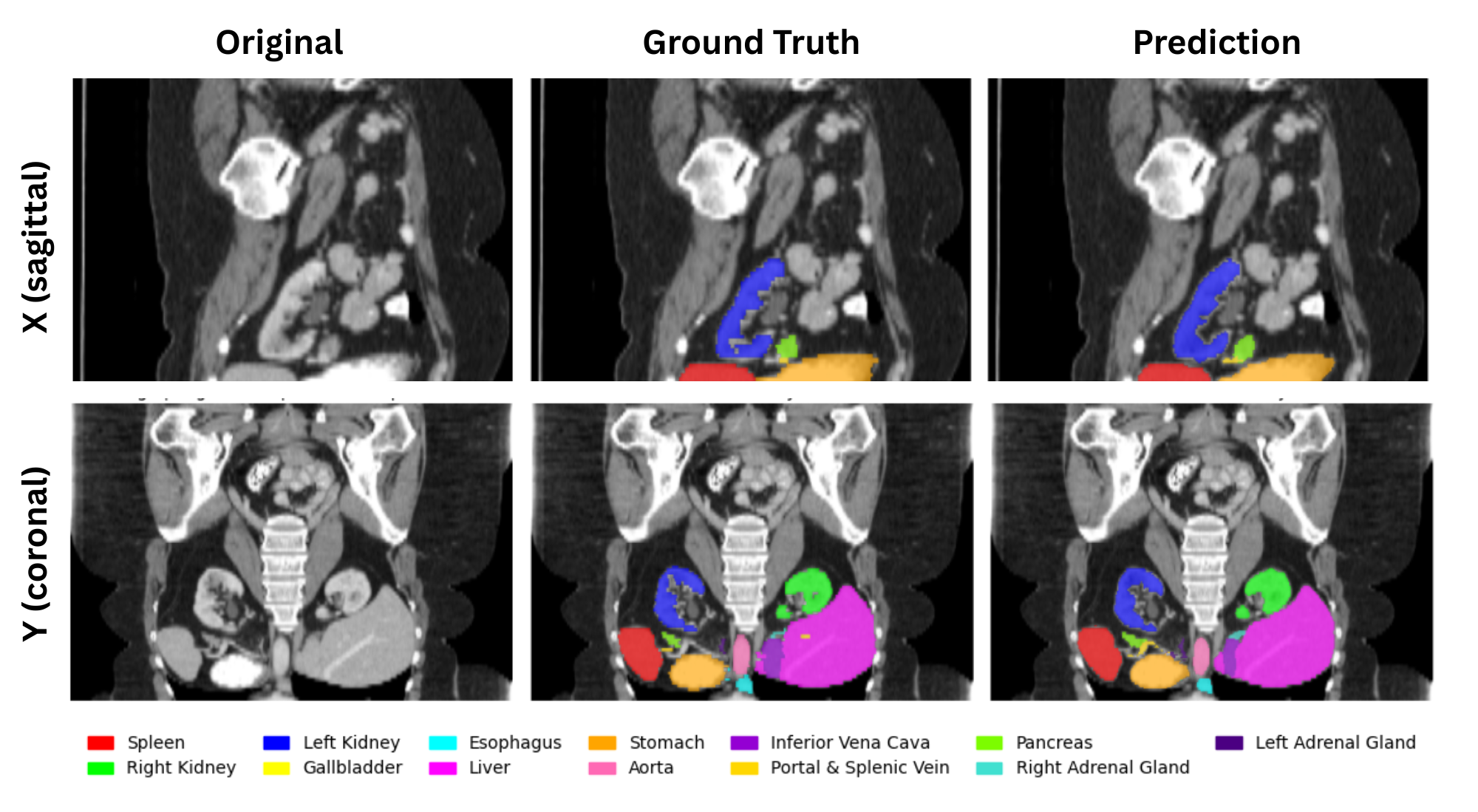}
    \caption{Qualitative results of SWINUNETR-RH on CT scans from the BTCV dataset. Each column shows a comparison between the original CT image, the ground truth segmentation, and the generated predictions.}
    \label{fig:swintext3d_results}
\end{figure}

The representative qualitative results of the proposed SwinTF3D framework are shown in Fig.~\ref{fig:swintext3d_results}. The figure illustrates how the model performs 3D abdominal organ segmentation on CT scans from the BTCV dataset. Each example presents the original CT slice, the corresponding ground truth segmentation, and the prediction generated by SwinTF3D. The close visual alignment between the predicted and reference masks demonstrates the accuracy and reliability of the proposed approach in capturing organ boundaries and anatomical structures across multiple organs.

\begin{table*}[t]
\centering
\caption{Organ-wise quantitative performance of the proposed SwinUNETR-RH model on the BTCV dataset. The table reports Dice Similarity Coefficient (DSC), Intersection over Union (IoU), F1-scores (F1\textsubscript{50}, F1, F2), Precision (P), Recall (R), and 95\% Hausdorff Distance (HD95). Average values across all organs are provided at the bottom.}
\label{tab:organwise_metrics}
\resizebox{\textwidth}{!}{
\begin{tabular}{lcccccccc}
\toprule
\textbf{Organ} & \textbf{DSC} & \textbf{IoU} & \textbf{F1\_50} & \textbf{F1} & \textbf{F2} & \textbf{Precision} & \textbf{Recall} & \textbf{HD95} \\
\midrule
Spleen                & 0.9558 & 0.9154 & 0.9573 & 0.9558 & 0.9545 & 0.9584 & 0.9537 & 2.593 \\
Right Kidney          & 0.9402 & 0.8876 & 0.9447 & 0.9402 & 0.9360 & 0.9479 & 0.9333 & 1.686 \\
Left Kidney           & 0.9433 & 0.8929 & 0.9441 & 0.9433 & 0.9425 & 0.9448 & 0.9420 & 1.844 \\
Gallbladder           & 0.6166 & 0.5105 & 0.6591 & 0.6166 & 0.5830 & 0.6945 & 0.5641 & 26.819 \\
Esophagus             & 0.7209 & 0.5679 & 0.7692 & 0.7209 & 0.6854 & 0.8110 & 0.6673 & 5.618 \\
Liver                 & 0.9659 & 0.9341 & 0.9674 & 0.9659 & 0.9645 & 0.9684 & 0.9637 & 7.710 \\
Stomach               & 0.8038 & 0.6888 & 0.8496 & 0.8038 & 0.7716 & 0.8929 & 0.7550 & 19.985 \\
Aorta                 & 0.8946 & 0.8113 & 0.9098 & 0.8946 & 0.8803 & 0.9206 & 0.8713 & 3.921 \\
Inferior Vena Cava    & 0.8342 & 0.7168 & 0.8483 & 0.8342 & 0.8232 & 0.8595 & 0.8174 & 5.093 \\
Portal \& Splenic Vein& 0.7155 & 0.5592 & 0.7348 & 0.7155 & 0.6993 & 0.7497 & 0.6901 & 9.564 \\
Pancreas              & 0.7828 & 0.6487 & 0.7927 & 0.7828 & 0.7750 & 0.8004 & 0.7709 & 5.345 \\
Right Adrenal Gland   & 0.7037 & 0.5441 & 0.6751 & 0.7037 & 0.7381 & 0.6586 & 0.7655 & 2.869 \\
Left Adrenal Gland    & 0.6544 & 0.4961 & 0.6402 & 0.6544 & 0.6743 & 0.6332 & 0.6918 & 5.428 \\
\midrule
\textbf{Average (per organ)} & \textbf{0.8101} & \textbf{0.7057} & \textbf{0.8225} & \textbf{0.8101} & \textbf{0.8021} & \textbf{0.8338} & \textbf{0.7989} & \textbf{7.325} \\
\bottomrule
\end{tabular}}
\end{table*}

\subsubsection{Comparison with State-of-the-Art Models on BTCV}

Table~\ref{tab:btcv-dice} presents the comparative Dice scores between the proposed SwinUNETR-RH and existing state-of-the-art methods on the BTCV dataset. The SwinUNETR-RH model demonstrates competitive performance, achieving an average Dice score of 81.01\%, which is comparable to advanced transformer-based models such as nnFormer (81.62\%) and UNETR++ (83.28\%). Notably, it outperforms several baselines including the original Swin-UNETR (80.44\%) and UNETR (76.00\%), indicating that the added refinement head effectively improves segmentation precision without requiring extensive retraining \cite{Isensee2024, Zhou2023, Huang2021}.

\begin{table*}[t]
\centering
\caption{Comparison of organ-wise Dice (\%) on the BTCV dataset between SwinUNETR-RH and other state-of-the-art models. The best results per column are highlighted in bold. These results were obtained from BTCV leaderboard }
\label{tab:btcv-dice}
\resizebox{\textwidth}{!}{
\begin{tabular}{lcccccccccccccc}
\toprule
Methods & Spl & RKid & LKid & Gal & Eso & Liv & Sto & Aor & IVC & PSV & Pan & RAG & LAG & Avg \\
\midrule
nnUNet       & \textbf{95.95} & 88.35 & 93.02 & 70.13 & 76.72 & 96.51 & \textbf{86.79} & 88.93 & 82.89 & \textbf{78.51} & \textbf{79.60} & \textbf{73.26} & \textbf{68.35} & 83.16 \\
TransBTS     & 94.55 & 89.20 & 90.97 & 68.38 & 75.61 & 96.44 & 83.52 & 88.55 & 82.48 & 74.21 & 76.02 & 67.23 & 67.03 & 81.31 \\
UNETR       & 90.48 & 82.51 & 86.05 & 58.23 & 71.21 & 94.64 & 72.06 & 86.57 & 76.51 & 70.37 & 66.06 & 66.25 & 63.04 & 76.00 \\
Swin-UNETR   & 94.59 & 88.97 & 92.39 & 65.37 & 75.43 & 95.61 & 75.57 & 88.28 & 81.61 & 76.30 & 74.52 & 68.23 & 66.02 & 80.44 \\
nnFormer     & 94.58 & 88.62 & 93.68 & 65.29 & 76.22 & 96.17 & 83.59 & 89.09 & 80.80 & 75.97 & 77.87 & 70.20 & 66.05 & 81.62 \\
UNETR++           & 94.94 & 91.90 & 93.62 & \textbf{70.75} & \textbf{77.18} & 95.95 & 85.15 & 89.28 & 83.14 & 76.91 & 77.42 & 72.56 & 68.17 & \textbf{83.28} \\
\midrule
\textbf{SwinUNETR-RH (Ours)}     & 95.58 & \textbf{94.02} & \textbf{94.33} & 61.65 & 72.08 & \textbf{96.59} & 80.38 & \textbf{89.46} & \textbf{83.43} & 71.55 & 78.29 & 70.36 & 65.45 & 81.01 \\
\bottomrule
\end{tabular}}
\end{table*}

\begin{table*}[!htbp]
\centering
\caption{Organ-wise Dice (\%) comparison across different methods on the Synapse multi-organ CT dataset. Abbreviations: Spl (Spleen), RKid (Right Kidney), LKid (Left Kidney), Gal (Gallbladder), Liv (Liver), Sto (Stomach), Aor (Aorta), Pan (Pancreas).}
\label{tab:organ-dice}
\resizebox{\textwidth}{!}{
\begin{tabular}{lccccccccc}
\toprule
\textbf{Methods} & \textbf{Spl} & \textbf{RKid} & \textbf{LKid} & \textbf{Gal} & \textbf{Liv} & \textbf{Sto} & \textbf{Aor} & \textbf{Pan} & \textbf{Avg DSC} \\
\midrule
U-Net          & 86.67 & 68.60 & 77.77 & 69.72 & 93.43 & 75.58 & 89.07 & 53.98 & 76.85 \\
TransUNet       & 88.91 & 85.08 & 77.02 & 81.87 & 94.08 & 75.62 & 87.23 & 55.86 & 77.49 \\
Swin-UNet       & 90.66 & 79.61 & 83.28 & 66.53 & 94.29 & 76.60 & 85.47 & 56.58 & 79.13 \\
UNETR         & 85.00 & 84.52 & 85.60 & 56.30 & 94.57 & 70.46 & 89.80 & 60.47 & 78.35 \\
MISSFormer     & 91.92 & 82.00 & 85.21 & 68.65 & 94.41 & 80.81 & 86.99 & 65.67 & 81.96 \\
Swin-UNETR     & 95.37 & 86.26 & 86.99 & 66.54 & 95.72 & 77.01 & 91.12 & 68.80 & 83.48 \\
nnFormer       & 90.51 & 86.25 & 86.57 & 70.17 & \textbf{96.84} & \textbf{86.83} & 92.04 & \textbf{83.35} & 86.57 \\
UNETR++             & \textbf{95.77} & 87.18 & 87.54 & \textbf{71.25} & 96.42 & 86.01 & 92.52 & 81.10 & \textbf{87.22} \\
\midrule
\textbf{SwinUNETR-RH (Ours)}       & 95.28 & \textbf{89.74} & \textbf{88.16} & 63.45 & 96.81 & 82.74 & \textbf{93.47} & 79.39 & 86.13 \\
\bottomrule
\end{tabular}}
\end{table*}

\begin{figure*}[t]
    \centering
    \includegraphics[width=\textwidth]{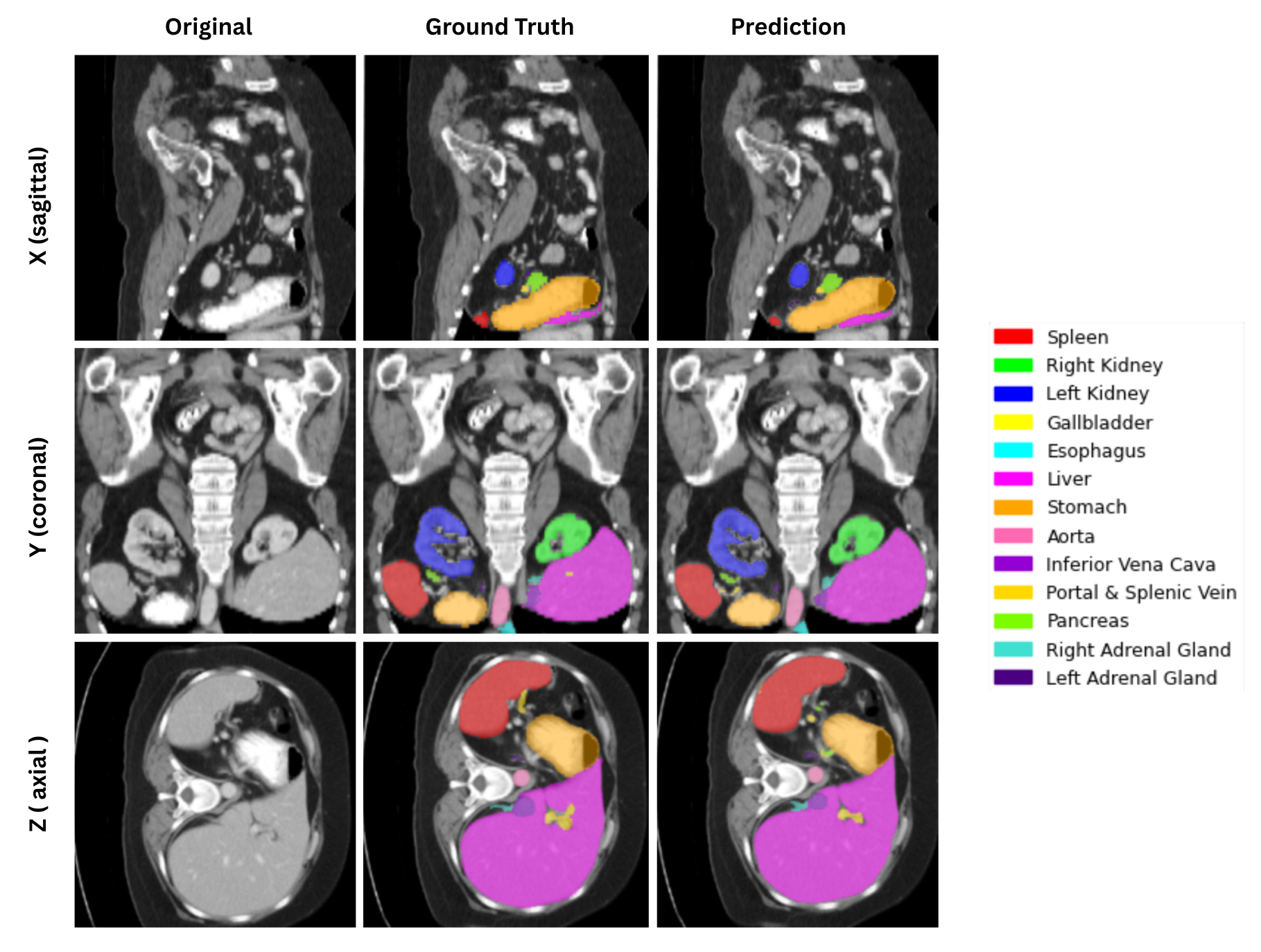}
    \caption{Qualitative visualization of 3D segmentation results in X (sagittal), Y (coronal), and Z (axial) views for CT volumes from the BTCV dataset. Each row shows the original CT image, the ground truth segmentation, and the corresponding prediction. The high visual similarity across all planes confirms the accuracy and spatial consistency of the proposed model.}
    \label{fig:xyz_views_results}
\end{figure*}

The model achieves particularly high accuracy for larger organs such as the spleen (95.58\%), right kidney (94.02\%), and left kidney (94.33\%), which are consistent with the organ-wise trends observed in Table~\ref{tab:organwise_metrics}. The results confirm that fine-tuning Swin-UNETR with the refinement head leads to enhanced spatial and structural consistency, contributing to overall performance gains. Fig.~\ref{fig:xyz_views_results} presents the qualitative visualization of 3D organ segmentation results obtained using the proposed SwinTF3D model. 

\begin{figure*}[t]
    \centering
    \includegraphics[width=\textwidth]{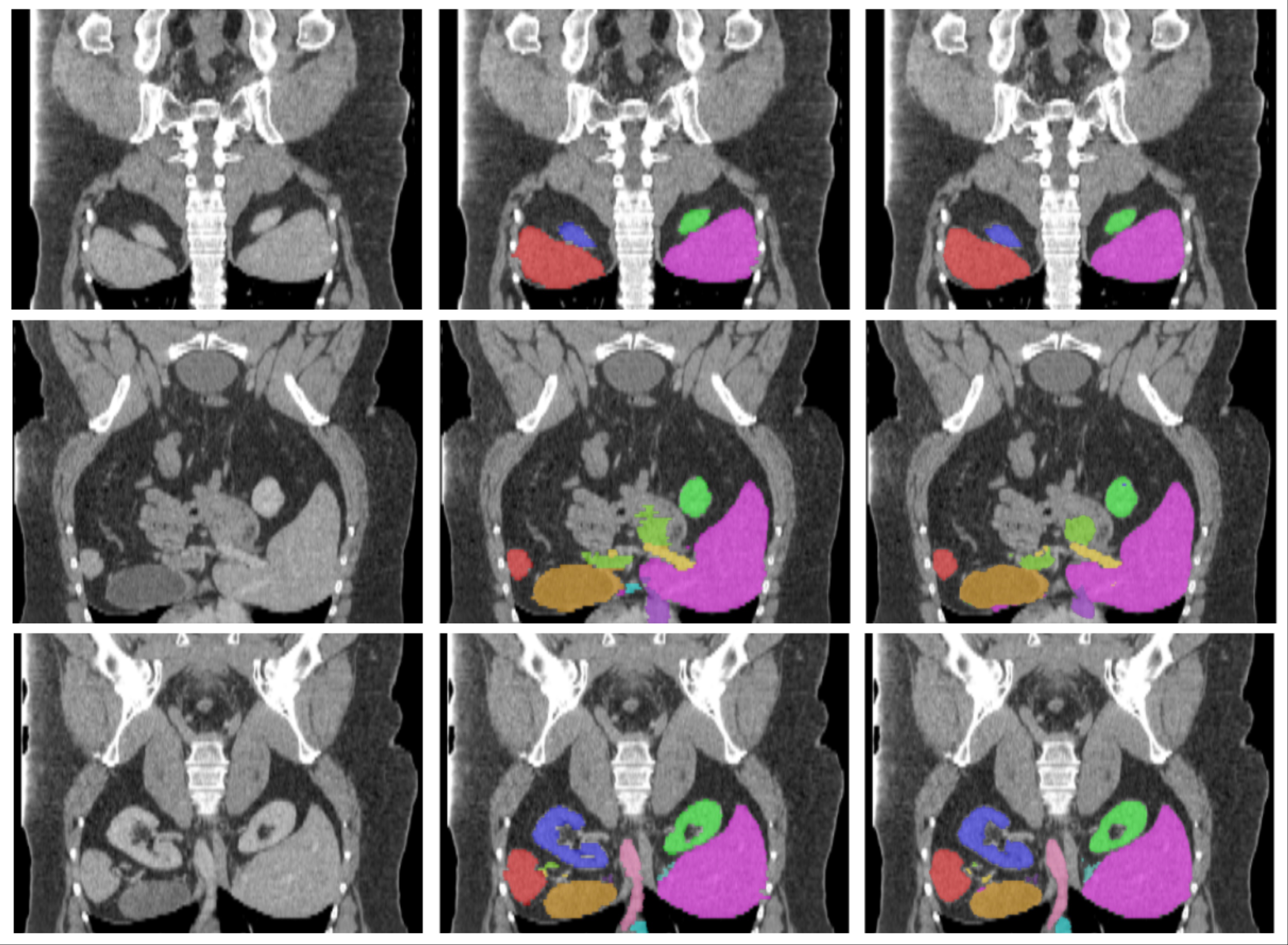}
    \caption{Visualization of segmentation results across multiple CT scan slices. Each row shows a different slice from the same volume, illustrating consistent organ delineation between the original image, the ground truth annotation, and the predicted segmentation produced by the proposed SwinTF3D framework.}
    \label{fig:multi_slice_ct_results}
\end{figure*}

We examine the performance of the suggested segmentation framework over several consecutive slices of volumetric CT data in order to further evaluate its robustness and spatial consistency. Visual inspection across several planes and slice indices becomes crucial because accurate 3D segmentation necessitates coherent predictions on isolated slices as well as throughout the entire volume. It is possible to assess whether the model preserves organ boundaries, maintains anatomical continuity, and avoids slice-wise inconsistencies that frequently occur in volumetric segmentation tasks by looking at segmentation outputs over a series of slices. The suggested method yields stable and anatomically consistent segmentations across various slices, enabling a thorough visual verification of segmentation quality throughout the volume, as shown by the qualitative results shown in Fig.~\ref{fig:multi_slice_ct_results}.

\subsubsection{Cross-Dataset Evaluation on Synapse}

To assess the generalization capability of the proposed SwinUNETR-RH model, we evaluated it on the Synapse multi-organ CT dataset without any additional fine-tuning. The Synapse dataset contains 8 annotated abdominal organs and differs from BTCV in imaging resolution, field of view, and label distribution. As shown in Table~\ref{tab:organ-dice}, SwinUNETR-RH achieves competitive Dice scores, demonstrating its ability to generalize across datasets with varying acquisition protocols. The model attains an average Dice score of 86.13\%, closely aligning with top-performing transformer architectures such as UNETR++ (87.22\%) and nnFormer (86.57\%). Consistent with the BTCV results, the proposed model performs strongly on larger organs like the liver, spleen, and aorta, while maintaining robust segmentation on smaller or irregular structures such as the pancreas and gallbladder. This reinforces the model’s capability to transfer learned anatomical representations across different medical imaging domains.
Overall, the results across BTCV and Synapse confirm that the fine-tuned SwinUNETR-RH model achieves consistent performance improvements over strong transformer-based baselines while maintaining excellent generalization across datasets.

\subsubsection{Prompt-Guided Inference Results on BTCV}

To further evaluate the effectiveness of the proposed SwinTF3D framework, we performed prompt-guided inference on the BTCV validation set using the synthetic prompt corpus described in Section~4.1.3. Each prompt referenced one or more target organs, and in some cases, included relational phrases such as “the region around the kidney that belongs to the liver.” The corresponding results are summarized in Table~\ref{tab:organwise_metrics_prompt}, which presents the quantitative metrics obtained across 13 abdominal organs under text-conditioned segmentation. The SwinTF3D model achieves an average Dice Similarity Coefficient (DSC) of 0.7745 and a mean Intersection over Union (IoU) of 0.6738 across all organs, demonstrating strong alignment between textual intent and volumetric segmentation. The F1-scores remain consistent across different thresholds (F1\textsubscript{50}, F1, F2), indicating balanced performance between precision and recall.

\begin{table*}[t]
\centering
\caption{Organ-wise quantitative performance of the proposed SwinTF3D model under prompt-guided inference on the BTCV validation set. The table reports Dice Similarity Coefficient (DSC), Intersection over Union (IoU), F1-scores (F1\textsubscript{50}, F1, F2), Precision (P), Recall (R), and 95\% Hausdorff Distance (HD95). Average values across all organs are shown at the bottom.}
\label{tab:organwise_metrics_prompt}
\resizebox{\textwidth}{!}{
\begin{tabular}{lcccccccc}
\toprule
\textbf{Organ} & \textbf{DSC} & \textbf{IoU} & \textbf{F1\_50} & \textbf{F1} & \textbf{F2} & \textbf{Precision} & \textbf{Recall} & \textbf{HD95} \\
\midrule
Spleen                & 0.9256 & 0.8793 & 0.9174 & 0.9258 & 0.9139 & 0.9194 & 0.9141 & 2.6280 \\
Right Kidney          & 0.9033 & 0.8568 & 0.9045 & 0.9037 & 0.8946 & 0.9029 & 0.8941 & 1.7290 \\
Left Kidney           & 0.9062 & 0.8645 & 0.9073 & 0.9060 & 0.8958 & 0.9064 & 0.8983 & 1.8710 \\
Gallbladder           & 0.5897 & 0.4703 & 0.6132 & 0.5924 & 0.5615 & 0.6535 & 0.5326 & 26.8460 \\
Esophagus             & 0.6911 & 0.5317 & 0.7322 & 0.6903 & 0.6522 & 0.7699 & 0.6289 & 5.6470 \\
Liver                 & 0.9325 & 0.8924 & 0.9379 & 0.9322 & 0.9268 & 0.9358 & 0.9239 & 7.7460 \\
Stomach               & 0.7746 & 0.6532 & 0.8028 & 0.7744 & 0.7441 & 0.8547 & 0.7203 & 20.0290 \\
Aorta                 & 0.8643 & 0.7812 & 0.8718 & 0.8641 & 0.8519 & 0.8744 & 0.8427 & 3.9540 \\
Inferior Vena Cava    & 0.8015 & 0.6849 & 0.8168 & 0.8024 & 0.7835 & 0.8236 & 0.7814 & 5.1360 \\
Portal \& Splenic Vein& 0.6838 & 0.5213 & 0.6989 & 0.6851 & 0.6627 & 0.7125 & 0.6573 & 9.5980 \\
Pancreas              & 0.7524 & 0.6129 & 0.7617 & 0.7521 & 0.7388 & 0.7642 & 0.7438 & 5.3960 \\
Right Adrenal Gland   & 0.6741 & 0.5145 & 0.6528 & 0.6749 & 0.6942 & 0.6335 & 0.7254 & 2.9100 \\
Left Adrenal Gland    & 0.6219 & 0.4648 & 0.6011 & 0.6236 & 0.6384 & 0.6047 & 0.6578 & 5.4700 \\
\midrule
\textbf{Average (per organ)} & \textbf{0.7745} & \textbf{0.6738} & \textbf{0.7891} & \textbf{0.7746} & \textbf{0.7623} & \textbf{0.7956} & \textbf{0.7537} & \textbf{7.3610} \\
\bottomrule
\end{tabular}}
\end{table*}

\begin{figure}[t]
    \centering
    \includegraphics[width=\columnwidth]{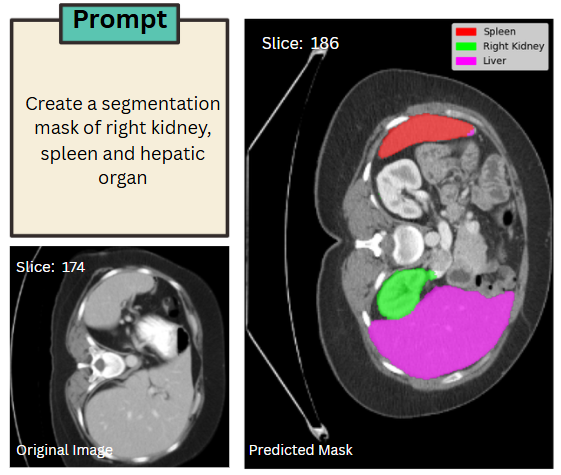}
    \caption{Example of text-guided segmentation using the proposed SwinTF3D framework. The model takes a natural language prompt specifying the organs of interest (right kidney, spleen, and liver) and produces accurate segmentation masks directly on the CT image, illustrating the model’s prompt-based controllability.}
    \label{fig:prompt_segmentation}
\end{figure}

Compared to the fine-tuned SwinUNETR-RH, the prompt-conditioned SwinTF3D demonstrates a slight reduction in overlap-based metrics, which is expected given the introduction of linguistic conditioning and the use of synthetic prompts rather than fixed label mappings. However, the model successfully captures prompt-specific contextual relationships, yielding interpretable and anatomically consistent segmentations. For instance, prompts such as “Create a segmentation mask of right kidney, spleen, and hepatic organ” demonstrate the model’s ability to correctly interpret organ references even when different terminologies or synonyms are used \cite{Bourigault2025}. The framework accurately identifies that the term “hepatic organ” refers to the liver and successfully generates coherent multi-organ segmentation masks, as visually illustrated in Fig.~\ref{fig:prompt_segmentation}. This highlights the SwinTF3D model’s capability to understand semantic variations within textual prompts and effectively capture spatial and anatomical relationships.

The results highlight that SwinTF3D maintains robust segmentation capability while extending the classical 3D segmentation paradigm to incorporate natural language interaction. This demonstrates the feasibility of prompt-driven organ segmentation even on datasets originally lacking textual annotations, opening new directions for multimodal learning in medical imaging.

\section{Discussion}
The experimental results across both visual-only and multimodal settings demonstrate the adaptability and efficiency of the proposed SwinTF3D framework. Fine-tuning the Swin-UNETR backbone with a lightweight refinement head enabled the visual encoder (SwinUNETR-RH) to achieve strong organ-level segmentation performance, outperforming or closely matching several state-of-the-art 3D transformer-based models. The refinement head improved boundary localization and inference stability, particularly for small or irregular organs such as the adrenal glands and pancreas, showing that targeted architectural refinement can yield meaningful gains without extensive retraining or large-scale data augmentation \cite{Cao2021,Chen2023}.

From a multimodal perspective, SwinTF3D provides an initial step toward integrating textual prompts into volumetric medical segmentation. The model successfully interprets natural language instructions and generates anatomically coherent segmentation maps aligned with textual intent. Although prompt-guided inference resulted in slightly lower Dice and IoU scores compared to the purely visual model, the performance drop was limited relative to the added semantic flexibility. These results indicate that textual embeddings can effectively modulate 3D visual representations, even when trained using synthetic prompts.

The relation-aware spatial priors incorporated in SwinTF3D further enhanced anatomical consistency between related structures. Prompts describing spatial dependencies, such as regions surrounding specific organs, produced smoother and more contextually aligned segmentations. This suggests that combining geometric and linguistic priors can improve interpretability without requiring large multimodal datasets or complex decoders. Instead, efficient fusion strategies paired with strong pretrained encoders can enable meaningful multimodal behavior under limited supervision.

Despite these advantages, certain limitations remain. The BTCV and Synapse datasets lack native textual annotations, requiring reliance on synthetically generated prompts. While effective for proof-of-concept validation, real clinical narratives may introduce richer linguistic structures that further enhance multimodal alignment. In addition, the current framework employs a single text encoder (BioClinicalBERT); exploring domain-specific language models trained on radiology or surgical text may further improve text–vision correspondence. Future work could also extend SwinTF3D to multimodal temporal data, such as dynamic MRI or ultrasound sequences \cite{sounack2025, chang2024emnetefficientchannel}. Overall, these findings highlight that combining pretrained volumetric encoders with lightweight text fusion heads offers a scalable and computationally efficient pathway toward prompt-driven 3D medical image understanding.

\section{Conclusion}

This study presented SwinTF3D, a multimodal framework for text-guided 3D medical image segmentation that combines a fine-tuned visual encoder (SwinUNETR-RH) with a lightweight fusion mechanism for linguistic conditioning. The proposed framework bridges the gap between conventional volumetric segmentation and emerging multimodal AI systems by enabling segmentation directly from natural language prompts. Comprehensive experiments on the BTCV and Synapse datasets demonstrated that the fine-tuned SwinUNETR-RH achieves high segmentation accuracy comparable to state-of-the-art transformer architectures while maintaining computational efficiency. Furthermore, SwinTF3D extends this capability to prompt-based inference, achieving stable performance and generating anatomically consistent masks guided by textual descriptions.

In future work, incorporating real-world clinical text, radiology reports, or procedural notes could enable more natural multimodal supervision. Extending SwinTF3D to handle 3D temporal data and cross-modality imaging (CT–MRI fusion) represents another promising direction \cite{liang2023, Hu2025}. Beyond these, future developments could focus on expanding the linguistic diversity and contextual understanding of the model by leveraging large-scale biomedical language models trained on domain-specific corpora. This would allow SwinTF3D to interpret complex clinical narratives and nuanced anatomical descriptions more effectively, improving its capacity to generalize across different clinical tasks and imaging modalities. In summary, this work provides evidence that prompt-based multimodal understanding is feasible in medical imaging and paves the way toward interactive, explainable, and language-aware 3D segmentation systems.

\bibliographystyle{IEEEtran}
\bibliography{references}

\end{document}